\documentclass{article} 
\usepackage{iclr2024_conference,times}

\usepackage{amsmath,amsfonts,bm}

\def\eqref#1{equation~\ref{#1}}

\def\1{\bm{1}}

\DeclareMathAlphabet{\mathsfit}{\encodingdefault}{\sfdefault}{m}{sl}
\SetMathAlphabet{\mathsfit}{bold}{\encodingdefault}{\sfdefault}{bx}{n}

\usepackage{hyperref}
\usepackage{url}
\usepackage{CJKutf8}

\usepackage{listings}
\usepackage{multirow}

\usepackage{graphicx}
\usepackage{subfigure}

\usepackage{times}
\usepackage{latexsym}
\usepackage{graphicx}
\usepackage{subfigure}
\usepackage{float}
\graphicspath{{images}}
\usepackage{stfloats}
\usepackage{booktabs}
\usepackage{indentfirst}
\usepackage{amsmath}
\usepackage{amssymb}
\usepackage{enumitem}

\usepackage{colortbl}
\usepackage{arydshln}

\title{PromptCBLUE: A Chinese Prompt Tuning Benchmark for the Medical Domain}

\author{Wei Zhu$^\ast$, \ \ Xiaoling Wang\thanks{Equal contributions. }, \ \ Huanran Zheng,   \\
East China Normal University \\
\AND
Mosha Chen\thanks{Corresponding author: chenmosha@holoflow.cn. Work done at Alibaba Group.}  \\
Holoflow Digital Technology, Hangzhou  \\
\And
Buzhou Tang \\
Harbin Institute of Technology, Shenzhen \\
}

\iclrfinalcopy 
\begin{document}

\begin{CJK*}{UTF8}{gbsn}

\maketitle

\begin{abstract}

Biomedical language understanding benchmarks are the driving forces for artificial intelligence applications with large language model (LLM) back-ends. However, most current benchmarks: (a) are limited to English which makes it challenging to replicate many of the successes in English for other languages, or (b) focus on knowledge probing of LLMs and neglect to evaluate how LLMs apply these knowledge to perform on a wide range of bio-medical tasks, or (c) have become a publicly available corpus and are leaked to LLMs during pre-training. To facilitate the research in medical LLMs, we re-build the Chinese Biomedical Language Understanding Evaluation (CBLUE) benchmark into a large scale prompt-tuning benchmark, PromptCBLUE. Our benchmark is a suitable test-bed and an online platform for evaluating Chinese LLMs' multi-task capabilities on a wide range bio-medical tasks including medical entity recognition, medical text classification, medical natural language inference, medical dialogue understanding and medical content/dialogue generation. To establish evaluation on these tasks, we have experimented and report the results with the current 9 Chinese LLMs fine-tuned with differtent fine-tuning techniques.\footnote{Codes are available at \url{https://github.com/michael-wzhu/PromptCBLUE}.} Our benchmark is released at \url{https://tianchi.aliyun.com/competition/entrance/532084/introduction}.

\end{abstract}

\section{Introduction}

Soon at its release, ChatGPT has taken all over the world by surprise. There are many large language models released since then, such as GPT-4 \citep{gpt4}, BARD\footnote{\url{https://bard.google.com/?utm_source=sem&utm_medium=paid-media&utm_campaign=q4enUS_sem6}}, ChatGLM\footnote{https://chatglm.cn/detail}, ERNIE Bot\footnote{https://yiyan.baidu.com/welcome}. Besides their ability to chat with humans, the research community is amazed by how much knowledge LLMs contains and how well they are at following instructions to complete a wide range of tasks without any additional model architectural designs. The open-sourced community has also seen a roaring growth in open-sourced LLMs, such as BLOOM\footnote{https://huggingface.co/bigscience}, LlaMA/LlaMA-2 \citep{Touvron2023Llama2O}, Falcon\footnote{https://huggingface.co/tiiuae}, Alpaca \citep{alpaca}, ChatGLM-6B\footnote{https://huggingface.co/THUDM/chatglm-6b}, Baichuan\footnote{https://github.com/baichuan-inc}, etc. With the fast advancement of open-domain LLMs, there are also many works trying to develop LLMs that are specialized in the medical domain, such as ChatMed\footnote{https://github.com/michael-wzhu/ChatMed}, ShenNong-TCM-LLM\footnote{https://github.com/michael-wzhu/ShenNong-TCM-LLM}, Ziya\footnote{https://huggingface.co/shibing624/ziya-llama-13b-medical-lora}. These models are playing a central role in bio-medical applications such as automatic health management of chronic patients, online diagnosis and consultations, and electronic health record understanding for insurance claims, etc. 

Evaluation datasets and benchmarks is one of the key recipe for developing high quality LLMs in the medical domain. In the pre-LLM era, there are many bio-medical benchmark datasets, including PubMedQA \citep{Jin2019PubMedQAAD}, BioRead \citep{pappas-etal-2018-bioread}, BLURB \citep{Gu2020DomainSpecificLM}, MedQA-USMLE \citep{Jin2020WhatDD}, Text2DT \citep{Text2dt} and the Chinese Biomedical Language Understanding Evaluation (CBLUE) benchmark. Recently, a series of medical LLM bechmarks are established, including \cite{gao2023dr}, ShenNong-TCM-Eval \citep{yue2023-TCMEB}. By analyzing the existing benchmark, we find that the existing benchmarks: (a) are limited to English which makes it challenging to replicate many of the successes in English for other languages, or (b) focus on knowledge probing of LLMs and neglect to evaluate how LLMs apply these knowledge to perform on a wide range of bio-medical natural language processing (BioNLP) tasks, or (c) have become a publicly available corpus and are leaked to LLMs during pre-training, such as MedQA-USMLE.

In order to address the above issues and facilitate medical LLM studies in Chinese, we re-build the CBLUE benchmark into PromptCBLUE, a large-scale multi-task prompt tuning benchmark dataset. PromptCBLUE is a suitable test-bed for Chinese medical LLMs for the following reasons: (a) we evaluate Chinese LLMs' multi-task capabilities on a wide range bio-medical tasks including medical entity recognition, medical text classification, medical natural language inference, medical dialogue understanding and medical content/dialogue generation. (b) PromptCBLUE's test ground truths are kept from the public corpus, which can effectively evaluate LLMs' generalization and instruction-following capabilities. (c) We host the dataset on an online platform for test evaluations and leaderboard presentation. 

To estable baseline evaluation on these tasks, we have experimented and reported the results of the currently most powerful commercial and open-sourced models with different learning or fine-tuning techniques. The following take-aways can be made: (a) despite its powerfulness, the OpenAI's ChatGPT and GPT-4 still fall behind fine-tuned open-sourced LLMs with smaller scales, showing that fine-tuning is still necessary toward applications.\footnote{Recently, OpenAI releases the fine-tuning APIs for its LLMs, demonstrating the necessity of fine-tuning. } (b) chain-of-thought prompting and output format design can help LLMs perform better on medical information extraction tasks of PromptCBLUE. (c) Currently, we can not see how further-pretraining in the medical domain significantly improve the performance of LLMs, which is different with the observations in the BERT era \citep{zhu-etal-2021-pre,guo-etal-2021-global}. We also conduct case studies to analyze when LLMs fail. 

The PromptCBLUE benchmark is an on-going work, and we intend to expand the tasks and build a more comprehensive GLUE like open platform for the Chinese BioNLP community. Overall, the main contributions of this study are as follows:

\begin{itemize}
\item We propose the first prompted based Chinese biomedical language understanding benchmark, PromptCBLUE, on the basis of CBLUE benchmark. The proposed benchmark serves as a platform for the development of Chinese LLMs for the medical domain. We encourages new dataset contributions. 
\item We report a systematic evaluation of a series of Chinese LLMs and learning/fine-tuning techniques, in order to understand the challenges derived by PromptCBLUE. We release the source code of the baselines as a toolkit for future research purposes. 
\end{itemize}

\section{Related work}

\subsection{Benchmarks for LLMs}

Benchmarking is essential for the advancement of natural language processing (NLP), especially in the era (or post-era) of pre-trained language models (PLMs). Notably, pre-PLM era has witnessed the development of many NLP models for a specific NLP task like reading comprehension \citep{rajpurkar-etal-2016-squad}, natural language inference \citep{bowman-etal-2015-large,sadat-caragea-2022-scinli,Zhu2021AutoNLUAS,zhu-etal-2021-discovering}, named entity recognition (NER) \cite{tjong-kim-sang-de-meulder-2003-introduction,OntoNotes50}, relation extraction \citep{zhu-2021-autorc}, and other classic benchmarking tasks \citep{barnes-etal-2022-semeval,ACE-2005,zuo-etal-2022-continually,Zheng2023CandidateSF,gao2023f}. With the rise of LMs like BERT \citep{devlin-etal-2019-bert}, RoBERTa \citep{Liu2019RoBERTaAR}, ALBERT \citep{Lan2020ALBERTAL} and Electra \citep{Zeng2021AutomatedCO,guo-etal-2021-global}, the benchmarks become multi-task. The prominent examples are: (a) the GLUE benchmark \citep{Wang2018GLUEAM}, which consists of multiple types of tasks in the format of single-sentence classification and sentence-pair classification. (b) SuperGLUE \citep{Wang2019SuperGLUEAS}, which consists more challenging tasks and more types of tasks compared to the GLUE benckmark. 

With the development of larger pretrained models, many benchmarks have been conquered. Thus, various bench-marking datasets have been proposed to evaluate the capabilities of pretrained LMs, including NaturalQuestion \citep{kwiatkowski-etal-2019-natural}, MultiSpanQA \citep{li-etal-2022-multispanqa}, commonsense reasoning \citep{clark2018think,sakaguchi2021winogrande},  mathematical reasoning \citep{cobbe2021training} and code generation \citep{zheng2023codegeex}. 

With the widely spread of large language models (LLMs) like GPT-3 \citep{brown2020language}, PaLM \citep{chowdhery2022palm}, ChatGPT\footnote{https://chat.openai.com/auth/login}, GPT-4 \citep{gpt4}, BARD \footnote{https://blog.google/technology/ai/try-bard/}, CLAUDE \footnote{https://claude.ai/} and ChatGLM\footnote{https://chatglm.cn/}, LLMs perform quite strong in some tasks such as summarization, translation, question answering, code generation \citep{qin2023chatgpt}. Thus, it is in-appropriate to continue to use these established benchmarks for evaluations of LLMs. In order to comprehensively assess the capabilities of LLMs, massive multi-task benchmarks are proposed, including MMLU \citep{hendrycks2020measuring}, BBH \citep{srivastava2022beyond}. MMLU \citep{hendrycks2020measuring} includes real-world exam questions of different tasks from multiple domains. It is widely used in evaluating LLMs. However, All these benchmarks are primarily focused on the English language.

Recently, Many evaluation benchmarks are proposed to evaluate the LLMs on Chinese, the language with the highest number of speakers. AGIEval \citep{zhong2023agieval} uses Chinese standardized exams like college entrance exams, math competitions, and lawyer qualification tests to build a test set for Chinese LLMs. In comparison, C-Eval \citep{huang2023c} includes question at four different education levels (elementary school, middle school, high school and college). CMMLU \citep{li2023cmmlu} consists of many Chinese-specific question such as Chinese food culture, Chinese driving rule. Our work complements the literature by providing a large scale prompt learning benchmark for the Chinese biomedical NLP field.  

\subsection{Benchmarking LLMs in the bio-medical domain}

Recently, there are works investigating whether LLMs not specifically pre-trained on the medical domains have medical domain knowledge. For example, \cite{singhal2023large} exams how LLMs perform on a series of benchmark tasks in the medical domain. Similarly, \cite{gao2023dr} also tested the LLMs on a collection of tasks. Recently, \cite{singhal2023towards} exams whether the currently most powerful LLMs like GPT-4, BARD can achieve expert level medical question answering. However, these works have focused on English medical benchmark tasks, while the evaluations on Chinese are neglected. In addition, while it is important to investigate the ability of LLMs on playing the role of human doctors and performing diagnostic question answering, it is of practical value to investigate how LLMs can conduct the tasks of Chinese medical text processing and thus provide assistance to medical professionals.\footnote{https://www.fastcompany.com/90922526/er-doctor-ai-medicine} Our PromptCBLUE benchmark consists of various tasks like summarizing doctor-patient dialogues that can help improve the efficiency of doctors, thus it is of value to industrial usage.

\subsection{Benchmarking Chinese medical text processing}

With the advances of language models, the community of Chinese bio-medical NLP has seen rapid development. After the rise of BERT, a series of pre-trained models in the medical domain are pre-trained or further pre-trained. For example, \cite{guo-etal-2021-global} propose a novel pretraining technique and obtain SOTA performances on Chinese spelling error correction tasks. Other representative Chinese medical pre-trained models includes \cite{Cai2021EMBERTAP,wang2021building}. With the advances of Chinese medical PLMs, a large-scale, multi-task benchmark, CBLUE \citep{zhang-etal-2022-cblue}, is developed to provide a unified testbed for those PLMs. CBLUE consists of a wide range of biomedical language understanding tasks which are carefully annotated by human experts. However, the CBLUE datasets does not come with the instruction tuning format, making it inconvenient to evaluate this benchmark on LLMs. Our PromptCBLUE benchmark builds upon CBLUE and transforms all the CBLUE tasks to instruction-tuning formats via a large collection of prompt templates that are validated by human experts.

\section{PromptCBLUE Benchmark}
\label{sec:promptcblue}

In this section, we first explain our motivations in proposing our PromptCBLUE benchmark, and then we will elaborate on the compositions of this benchmark.

\subsection{Motivation}

\textbf{Probing the medical knowledge of LLMs} \quad Knowledge, the information encompassing facts, events, and skills, has long served as an central indicator for intelligence level of artificial intelligence (AI) \citep{yu2023kola}. A series of works have been devoted in investigating whether LLMs contains medical knowledge via large scale pretraining \citep{singhal2023large,gao2023dr,singhal2023towards}. However, these works have focused on English medical benchmark tasks, while the evaluations on Chinese are neglected. 

\textbf{Capabilities of multi-task medical text understanding} \quad Nowadays, the medical application requires a system to be good at multiple tasks at once \citep{Karabacak2023EmbracingLL}. For example, a system for medical query answering may requires the system to recognize the key entities in the query, or is able to retrieve semantically related known queries or related medical documents that could help to answer the queries, and summarize the medical knowledges, facts or observations to respond to the users. In the the era of LLMs, whether a single model can perform various tasks are widely investigated \citep{wei2021finetuned,naturalinstructions,supernaturalinstructions,chung2022scaling,ouyang2022training}. The multi-task or cross-task generalization abilities will re-shape the research and development processes in the industry.\footnote{https://www.linkedin.com/pulse/revolutionizing-seo-ai-embracing-chatgpt-4-llm-technology-business} PromptCBLUE contains a wide range of Chinese bio-medical tasks, thus providing a ideal testbed for developing and testing the capabilties of Chinese LLMs for medical applications.


\textbf{PromptSource for Chinese medical NLP community} \quad A large scale prompt set is of central importance for the LLMs' development. For example, PromptSource \citep{bach-etal-2022-promptsource}, Natural Instructions \citep{naturalinstructions}, and many other prompt tuning datasets play an important role in facilitating the instruction tuning of LLMs. In the Chinese bio-medical domain, the instruction tuning datasets are relatively lacking. Many instruction datasets are developed via self-instruct \citep{wang2022self}, there are is still no large scale instruction datasets for Chinese medical text understanding. Thus, in PromptCBLUE, we construct a large prompt tuning dataset for Chinese medical LLMs, which can facilitate and speedup the development and evaluation of Chinese medical LLMs.

\subsection{Overview of PromptCBLUE}

Built upon the CBLUE benchmark \citep{zhang-etal-2022-cblue}, we created an extensive multi-task test suite in the medical domain for the LLMs that supports Chinese. The tasks in PromptCBLUE can be divided into the following groups:

\begin{itemize}
\item \textbf{Medical information extraction}, including: (a) CMeEE-V2 \citep{Zan2020BuildingAP}, a medical named entity recognition task. (b) CMeIE \citep{Guan2020CMeIECA}, a medical triple extraction task. (c) CHIP-CDEE, which asks models to extract clinical finding events; (d) CHIP-CDN, which asks one to map the diagnosis descriptions to standard ICD-10\footnote{
https://www.whofic.nl/familie-van-internationale-classificaties/referentie-classificaties/icd-10} disease terms. (e) IMCS-V2-NER for extracting medical entities in the dialogues from the medical dialogue datasets IMCS-V2 \citep{Chen2022ABF}. 
\item \textbf{Medical text classification}, which includes: (a) CHIP-CTC \citep{Zong2021SemanticCO}, the classification task for Chinese eligibility criteria in clinical trials. (b) KUAKE-QIC, which classifies online medical queries. (c) IMCS-V2-DAC for medical intent classification of medical dialogues. 
\item \textbf{Medical natural language inference tasks}, including CHIP-STS \citep{CHIP-STS}, KUAKE-QQR, KUAKE-IR and KUAKE-QTR tasks, which asks a model to determine the semantic relations between a pair of medical queries or a query-document pair. 
\item \textbf{Symptom status understanding for medical dialogues}. Medical dialogues, like medical consultations, are centered around paitents' symptoms. However, not every symptom mentioned in the dialogues is related to patients or reflect patients' current medical status. Thus, in order to gain deep understanding of the doctor-patient dialogues, a model must be able to extract the symptoms and their statuses. This cohort includes: (a) IMCS-V2-SR, and (b) CHIP-MDCFNPC for extracting clinical findings and their status. This cohort of tasks is based on the medical dialogue datasets IMCS-V2 \citep{Chen2022ABF} and CHIP-MDCFNPC \citep{MDCFNPC}. 
\item \textbf{Medical content generation}, which includes two generation tasks based on medical dialogues between patients and doctors: (a) IMCS-V2-MRG \citep{Chen2022ABF}, which asks LLMs to summarize the medical dialogues. (b) MedDG \citep{Liu2020MedDGAL}, which asks LLMs to act like a doctor and respond to patient's queries in a dialogue. 
\end{itemize}

\textbf{Prompt collection} \quad We employ both manual efforts and ChatGPT to collect prompt templates for PromptCBLUE. The prompt templates mainly contains task instructions, that is, what we want the LLMs provide given the text input. Firstly, each of the three annotators who are graduates student majoring in computer science and are studying LLMs will write around 2 seed prompt templates manually. We ask the annotators to write the prompts as diversified as possible. The prompt templates are then reviewed by a review panel consisting of two medical experts and a senior NLP researcher to ensure their validity. If a prompt is not proper (for example, not expressing the task cleary), we will ask the annotators to modify it until it was accepted by the domain experts. Then, we will ask the ChatGPT to rephrase each of the six seed prompts templates for 10 times without changing the meaning or changing the placeholders in the templates. Then the generated templates will be reviewed by the same panel of experts, and only the templates passing the reviews will be added to the template pool. After the prompt collection process, they are totally  \footnote{The manually written prompt templates and the augmented template sets are open-sourced at our code repository, https://github.com/michael-wzhu/PromptCBLUE. } 

\textbf{Additional contents for prompts} \quad In addition to the above prompt templates, we may add the following contents in the prompts:
\begin{itemize}
    \item Explanations of the label set, that is, what each label means; 
    \item The specifications of the output formats. This is particularly important for us to process the output sequences of LLMs on information extraction tasks. 
    \item Demonstrations. This part is especially important for LLM services that we are unable to fine-tune. 
    \item Instructions that asks the LLMs to complete the task step-by-step or reason step-by-step. 
\end{itemize}

We will conduct ablation studies on how these parts of prompts will affect the performances of LLMs like ChatGPT.

\textbf{Response format} \quad Note that LLMs can only generate token sequences to represent their answers to the queries, we have to transform the structured outputs of the original CBLUE tasks to natural language sequences. In Appendix \ref{subsec:format}, we present the target output formats for each task under PromptCBLUE.

\textbf{Sample format} \quad We use the prompt template pool to reformulate all the tasks in the CBLUE benchmark to instruction-following generation tasks with the following data structure: 
\begin{verbatim}
{
  "input": str,
  "target": str,
  "answer_choices": Union[list, Nonetype],
  "sample_id": str,
  "task_type": str,
  "task_dataset": str,
}
\end{verbatim}
Here, \emph{input} is the prompt sequence, \emph{target} is the response we want the model to output, or at least the model should generate responses that contains the answers to the prompt. The other four keys are auxiliary, and they will not be used by the LLMs. \emph{sample\_id} is the sample index. \emph{answer\_choices} is the label options allowed by the prompt. The value for this key will be \emph{None} if the task does not have a predefined label set. \emph{task\_dataset} specifies the task name in the original CBLUE benchmark, and \emph{task\_type} is the original CBLUE task type.

\textbf{Dataset splits} \quad Note that after filling in the CBLUE data samples in the prompt templates, the train/dev/test sets for some task will be quite large. Considering that LLMs require high computation resources to fine-tune and have large latency for token sequence generation, we limit the training samples of each task to 3000 to 5000, and the dev/test set to 600-800. We first fill each prompt template with the samples to construct a large test sample pool and randomly select the prompt-response pairs via uniform sampling.\footnote{Note that the training set is provided just as a reference for participants in the Tianchi competition, since some of the original CBLUE tasks have large training sets.} In Table \ref{tab:stats} of Appendix \ref{subsec:stat}, we present the dataset statistics for the PromptCBLUE dataset.

\textbf{Quality Checking} \quad The quality check of our data are conducted in the following three aspects:
\begin{itemize}
\item Ensuring the quality of the prompt templates with the help of the expert panel, as described above.
\item Checking the quality of CBLUE benchmark. During the development of PromptCBLUE benchmark, we are also helping the CBLUE benchmark to improve the annotations. For example, we have found that the original QQR task has an in-proper label set, and ask the CBLUE organizers to re-annotate this task. 
\item Random sampling. To ensure the data quality, we sampled 5\% or 200 of the samples from each task of PromptCBLUE, and each sample is examined by a group of annotator from the medical field. Finally, we identify an average of 0.9\% mislabeling rate. Based on the evaluation results in the next section, we believe such an error rate will not significantly impact the overall evaluation accuracy.
    
\end{itemize}

\section{Experiments}

\subsection{Experimental settings}

We now conduct a comprehensive evaluation of a series of existing open-sourced LLMs or LLMs with publicly available APIs on medical knowledge and language understanding within the context of Chinese, via our PromptCBLUE dataset. We report their performances under different settings and investigate several factors that could affect the performance of LLMs on PromptCBLUE.

\noindent \textbf{Evaluation Setup} \quad Our evaluations are conducted under the following three settings:
\begin{itemize}
\item Few-shot in-context learning setting. Under this setting, we investigate the in-context learning ability \cite{Li2023UnifiedDR,Wang2023LearningTR} of LLMs by concatenating demonstrating instruction-response pairs from the training set to the current instruction. For tasks with label sets, we concatenate demonstrations containing at least $k$ instances for each label except the CHIP-CTC task.\footnote{In this task, since the size of the label set is large (45 labels), we concatenate the $k$ demonstrations with different labels. } For other tasks, we concatenate the $k$ demonstrations. The exact value of $k$ varies across tasks due to the different sequence length and label set sizes. The exact number of $k$ for each task is reported in Table \ref{tab:results_few_shot_in_context}.\footnote{Currently, we provide fixed demonstrations for all the test samples on a given task. In the future, we will explore how demonstration selections affect the performances. }

\item Few-shot fine-tuning learning setting. Note that the selected demonstrations from the previous setting could also be treated as training samples to fine-tune the open-sourced models. We also conduct LLM fine-tuning with different numbers of samples to investigate how training sample sizes affect the model test performances.

\item Fine-tuning. We will fine-tune a series of open-sourced LLMs on the large scale training set (introduced in Section \ref{sec:promptcblue}) and report their performances. In terms of the fine-tuning techniques, we will experiment both full-model fine-tuning and a series of parameter efficient fine-tuning (PEFT) methods. Note that during fine-tuning, we treat the entire PromptCBLUE as one task and use only one set of PEFT parameters for predictions of all the test samples. 


\end{itemize}

Under all the above settings, we ask LLMs to generate output and used a series of regular expressions to extract the model’s
choice.\footnote{The example scripts for processing the LLMs' outputs are provided at \url{https://github.com/michael-wzhu/PromptCBLUE/blob/main/src/evaluation/post_generate_process.py}}

\textbf{Evaluation metrics} \quad Since metrics like BLUE or ROUGE \cite{lin-2004-rouge} can not properly measure how LLMs perform for some of the PromptCBLUE tasks like medical information extraction tasks, we use post-processing scripts to transform the output sequences to structured data formats. PromptCBLUE adopt the following metrics:
\begin{itemize}
\item Instance-level strict micro-F1 for medical information extraction tasks, IMCS-V2-SR and CHIP-MDCFNPC. Here, an instance means a complete piece of information extracted from the given document. For example, in CMeEE-V2, an instance consists of a entity mention extracted and its predicted entity label. And in IMCS-V2-SR, an instance consists of two keys: the entity mention of a symptom, and its status. We adopt the strict metrics, meaning that the model predicts an instance correctly if and only if it correctly predicts the all the keys of an instance. 
\item For medical text classification tasks and the IMCS-V2-DAC tasks, we adopt the macro-F1 score. 
\item For medical natural language inference tasks, we adopt the micro-F1 score. 
\item For the medical content generation tasks, we adopt ROUGE-L \cite{lin-2004-rouge} as the metric. 
\end{itemize}

\noindent\textbf{Models} \quad We assess a collection of well-known language models in different sizes. For commercial models with publicly available APIs, we evaluate GPT-4 \cite{OpenAI2023GPT4TR}, ChatGPT\footnote{We evaluate ChatGPT and GPT-4 via the OpenAI APIs.}, and ChatGLM.\footnote{The evaluations for three commercial LLMs are conducted during July 9th to July 11th, 2023. } For open-sourced models, we consider Baichuan-13B as the main model. We also compare this LLM with the following (L)LMs: (a) GPT-2 Chinese\footnote{https://huggingface.co/uer/gpt2-chinese-cluecorpussmall}. (b) Randeng-T5-784M\footnote{https://huggingface.co/IDEA-CCNL/Randeng-T5-784M-MultiTask-Chinese}. (c) BLOOMZ-7.1B-mt\footnote{https://huggingface.co/bigscience/bloomz-7b1-mt}. (d) ChatGLM-6B-2. (e) ChatMed\footnote{https://github.com/michael-wzhu/ChatMed}, which is adopted from the LlaMA-7B backbone. (f) Chinese-LLaMA-2 7B/13B\footnote{https://github.com/michael-wzhu/Chinese-LlaMA2}, which are the Chinese version of LlaMA-2 models \cite{Touvron2023Llama2O} from Meta. (g) Ziya-13B-medical\footnote{https://huggingface.co/shibing624/ziya-llama-13b-medical-lora}, which is also further pre-trained with the LlaMA-2 models.

\noindent\textbf{Fine-tuning methods} \quad For fine-tuning the open-sourced LLMs, we do not consider full-model fine-tuning as in \citet{devlin-etal-2019-bert} due to resource limitation. For parameter efficient fine-tuning, we adopted the following methods: (a) P-tuning \cite{Liu2021GPTUT}. (b) P-tuning v2 \cite{liu-etal-2022-p}. (c) Parallel adapter by \cite{He2021TowardsAU}. (d) LoRA \cite{hu2021lora}. (e) AdaLoRA by \cite{Zhang2023AdaptiveBA}. We adjust the hyper-parameters of these methods so that the tunable parameters accounts for less than 1\% of the LLM backbones' parameters. The detailed hyper-parameter settings for each method are provided in Appendix \ref{subsec:hyperparams}. After a model is fine-tuned, we use beam search (beam size 4) for generating the responses.

\subsection{Main results}

\textbf{Results for the few-shot in-context learning setting} \quad 
Table \ref{tab:results_few_shot_in_context} shows the performance of ChatGPT, GPT-4 and ChatGLM under the few-shot setting. These three models are well-established LLMs with commercial APIs. 


From Table \ref{tab:results_few_shot_in_context}, we can see that OpenAI's GPT-4, achieve the best overall performances, and outperforms the Chinese LLM ChatGLM with the comparable model size. We believe this is because OpenAI has advantages in the large scale instruction tuning dataset collected from the real users and thus its models are better at following different instructions. 




\begin{table}[t]
\caption{The few-shot in-context learning results on the PromptCBLUE test set. }
\label{tab:results_few_shot_in_context}
\begin{center}
\begin{tabular}{c|c|ccc}
\hline
\textbf{Task}   &  \#Demos   &   ChatGPT   & GPT-4  &   ChatGLM     \\
\hline
\multicolumn{5}{c}{\textbf{\emph{Medical information extraction tasks}}}  \\
\hline
CMeEE-V2  &  7   &    0.469   &  0.572   &  0.380 \\
CMeIE   &  12  &   0.305  &  0.457   &  0.266 \\
CHIP-CDEE    &  5  &    0.283   &  0.375  &  0.182 \\
CHIP-CDN  &  10  &   0.606   &  0.856  &  0.477   \\ 
IMCS-V2-NER   &   10  &   0.568   &  0.695   &   0.549  \\

\hline
\multicolumn{5}{c}{\textbf{\emph{Medical text classification tasks}}}  \\
\hline
CHIP-CTC     &  10 &    0.525   & 0.614   & 0.431       \\
KUAKE-QIC    &  10   &   0.485    &   0.563   &   0.430  \\   
IMCS-V2-DAC   &    8    &   0.345    &  0.435   &    0.318   \\

\hline
\multicolumn{5}{c}{\textbf{\emph{Medical natural language inference tasks}}}  \\
\hline
CHIP-STS      &  10  &    0.710    &  0.728  &  0.631 \\
KUAKE-QQR    &  10  &   0.304   &  0.417  &   0.259   \\
KUAKE-IR     &     5    &    0.518   &   0.573  &  0.453   \\
KUAKE-QTR    &   10   &    0.231   &  0.364   &  0.218  \\

\hline
\multicolumn{5}{c}{\textbf{\emph{Symptom status understanding for medical dialogue}}}  \\
\hline
IMCS-V2-SR    &   5   &   0.330     &  0.395   &  0.283   \\
CHIP-MDCFNPC  &   3   &   0.585    &   0.673   &  0.512  \\

\hline
\multicolumn{5}{c}{\textbf{\emph{Medical content generation}}}  \\
\hline
IMCS-V2-MRG   &   2   &   0.325   &  0.373  & 0.303 \\
MedDG    &   5   &   0.136   &   0.193   &  0.108   \\

\hline 
\multicolumn{5}{c}{\textbf{\emph{Overall}}}  \\
\hline 
Overall score   &   -   &   0.421    &   0.518  &   0.436  \\

\hline
\end{tabular}
\end{center}
\end{table}

\textbf{Fine-tuned results} \quad We now fine-tune the open-sourced LLMs with the large-scale training set described in Table \ref{tab:stats}.

In Table \ref{tab:results_ft}, we first demonstrate how different fine-tuning methods affect the model performance, given the Chinese-LlaMA-2 13B backbone. The following take-aways can be made from Table \ref{tab:results_ft}: (a) when fine-tuning with the large-scale training set, the overall performance ranking of the parameter efficient tuning methods is: LoRA $>$ AdaLoRA $>$ Adapter $>$ P-tuning-v2 $>$ P-tuning, which is consistent with \cite{Ding2022DeltaTA}. (b) AdaLoRA does not perform better than LoRA under this setting. (c) After fine-tuning with the large scale training set, the open-sourced model, Baichuan 13b, has obtained an average score of 71.0\%, which is 19.2\% higher than GPT-4, demonstrating that fine-tuning is necessary for achieving high scores on various prompt-based tasks, despite the existence of the current powerful LLMs. (c) We can see that no fine-tuning method can obtain the best scores on all five task cohorts. This could mean that in real world applications, in order to achieve better performances, one can use different sets of PEFT parameters to deal with different tasks.

\begin{table}[t]
\caption{The fine-tuning results with the Baichuan 13B backbone. }
\label{tab:results_ft}
\begin{center}
\begin{tabular}{c|cccccc}
\hline
\textbf{Task}   &   P-tuning  &  P-tuning-v2  & Adapter  &    LoRA   & AdaLoRA     \\

\hline

Medical information extraction  &  0.732   &  0.728  &  0.742   &  0.749    &    0.751   \\
Medical text classification  &   0.814  &  0.811  &  0.834   &  0.832  &  0.826  \\
Medical natural language inference  &   0.742   &   0.759   & 
 0.768   &  0.775   &   0.769   \\
Symptom status understanding  &   0.705   &   0.713   &  0.739   &    0.736   &   0.738  \\
Medical content generation  &   0.236  &  0.255   &  0.261   &  0.277   &  0.273  \\
\hline
Overall score   &   0.685   &    0.690   &   0.705  &  0.710  &   0.708  \\
\hline 

\end{tabular}
\end{center}
\end{table} 


Second, Table \ref{tab:results_ft_different_size} reports how different LLMs backbones perform differently, after being fine-tuned with the LoRA method. From Table \ref{tab:results_ft_different_size}, we can observe that: (a) generally, the open-sourced models with 13B size perform better than those with 6B/7B parameters, more even fewer parameters. (b) despite being trained with medical corpus, ChatMed does not perform significantly better than Chinese LlaMA-2-7b, and Ziya-medical-13B does not perform significantly better than BaiChuan 13B, an open-domain LLM. This observation demonstrates that our PromptCBLUE benchmark does not only requires medcial knowledge, general language understanding and generation capabilities also play an equally important role.   

\begin{table}[t]
\caption{The fine-tuning results of different LLM backbones when it is fine-tuned with LoRA. }
\label{tab:results_ft_different_size}
\begin{center}
\begin{tabular}{c|cccccc}
\hline
\textbf{Model}  &   \textbf{Overall score}  \\
\hline
GPT-2 Chinese  &  0.213  \\
Randeng-T5-784M  &   0.294   \\
BLOOMZ-7.1B-mt  &    0.596  \\
ChatGLM-6B-2  &  0.681  \\
ChatMed   &   0.688   \\
Chinese-LlaMA-2 7B  &  0.683   \\
Chinese-LlaMA-2 13B  &   0.704  \\
Ziya-13B-medical   &   0.705   \\
Baichuan-13B  &   0.710  \\
\hline 

\end{tabular}
\end{center}
\end{table}

\subsection{Analysis}

\subsubsection{Error analysis} \quad We now conduct case studies and error analysis on the following tasks: CMeEE-V2, a sequence labeling task, KUAKE-QIC, a classification task, and IMCS-V2-MRG, a medical content generation task. 

On the test set of CMeEE-V2 task, with 400 test samples, BaiChuan 13b predicts 1996 entities. Among the 1996 entities predicted, 71.3\% are correct in both entity mentions and entity labels. Among the wrongfully predicted entities, 5.2\% are erroneous due to wrong labels, 11.4\% are due to wrong entity boundaries, and the rest 13.1\% are mentions that should not be an entity. Among the entity instances in the ground truth, 13.6\% are entity mentions that are entirely omitted by the LLM. Table \ref{tab:error_cmeee} presents three examples demonstrating the errors of the fine-tuned BaiChuan 13b. In the first example, the LLM fails to detect the clinical finding entity \emph{分泌物干燥} (dry discharge). In the second example, the LLM generate incomplete mentions for \emph{剪短病儿指甲} (Cut the nails of sick children short). In the third example, the LLM mistakenly generate \emph{结节} (nodules) and \emph{窦道} (sinus tracts) as body part entities.

\begin{table}[ht]
\caption{Error cases of the fine-tuned BaiChuan 13b on the CMeEE-V2 task. - means mentions that are omitted by the LLM or not in the ground truth.  }
\label{tab:error_cmeee}
\begin{center}
\begin{tabular}{p{6.5cm}p{1.4cm}p{1.2cm}p{1.4cm}p{1.2cm}}
\hline
\textbf{Input Sentence}  &   \multicolumn{2}{c}{\textbf{Ground truth}}  & \multicolumn{2}{c}{\textbf{Predicted}}   \\
&   \textbf{mention}   &  \textbf{type}    &   \textbf{mention}   &  \textbf{type}  \\
\hline

请从以下文本中提取医疗命名实体：$\backslash$n抗组胺药可使分泌物干燥，不宜服用。$\backslash$n医疗命名实体的类型包括：疾病，临床表现，身体部位，医疗程序，医院科室，药物，医学检验项目，微生物类，医疗设备 \quad (Please extract medical named entities from the following text: $\backslash$nAntihistamines can cause dry discharge and should not be taken. $\backslash$nTypes of medical named entities include: disease, clinical finding, body part, medical procedure, hospital department, drug, medical test item, microorganism, medical equipment)   &  分泌物干燥 (dry discharge)  &  临床表现 (clinical finding)  &  -  &   -   \\

\hline 



剪短病儿指甲避免搔破皮疹，继发细菌感染。$\backslash$n问题：请标出句子中的疾病，临床表现，身体部位，医疗程序，医院科室，药物，医学检验项目，微生物类，医疗设备实体。 \quad (Cut the nails of sick children short to avoid scratching the rash and causing secondary bacterial infection. $\backslash$nQuestion: Please mark the disease, clinical finding, body part, medical procedure, hospital department, drug, medical test item, microorganism, and medical equipment entity in the sentence.) &    剪短病儿指甲 (Cut the nails of sick children short)   &  医疗程序  (medical procedure)   &   剪短  (cut)   &   医疗程序 (medical procedure)  \\

& 搔破皮疹 (scratching the rash)  &   临床表现 (clinical finding)   &     皮疹 (rash)    &   临床表现 (clinical finding)   \\
\hline 

在下面的文本中找出医疗命名实体：$\backslash$n开始在局部皮肤形成皮下结节，以后结节软化、破溃形成窦道。$\backslash$n请从以下类型中选择医疗命名实体：疾病，临床表现，身体部位，医院科室，医学检验项目，微生物类，医疗设备 \quad (Find the medical named entity in the following text: $\backslash$n begins to form subcutaneous nodules in the local skin, and later the nodules soften and rupture to form sinus tracts. $\backslash$nPlease select medical named entities from the following types: disease, clinical finding, body part, hospital department, medical test item, microbiology, medical equipment)   &   -   &  -   &   结节 (nodules)  &   身体部位 (body part)  \\

&   -   &  -   &   窦道 (sinus tracts)   &   身体部位 (body part)   \\

\hline 

\end{tabular}
\end{center}
\end{table}

\begin{figure}[h]
\begin{center}
\includegraphics[width=0.66\textwidth]{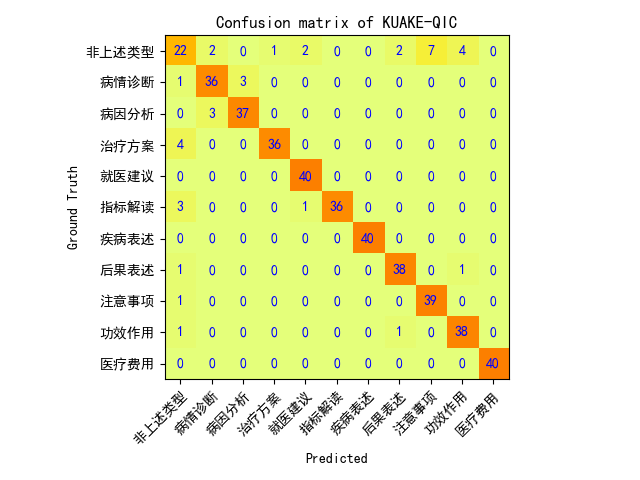}
\end{center}
\caption{The confusion matrix of the LLM on the KUAKE-QIC test set.}
\label{fig:confusion_matrix}
\end{figure}

Figure \ref{fig:confusion_matrix} presents the classification confusion matrix for LlaMA-2 13b on the test set of the KUAKE-QIC task. On the test set of KUAKE-QIC task, each label has 40 samples. Note that we discard the \emph{其他} (other) label in the original dataset, and if the input text is not in the given categories of the prompt, we will ask the LLM to generate the \emph{非上述类型} (not in the given categories) label. Thus the samples with the \emph{非上述类型} label will be either from the samples with the \emph{其他} (other) label, or in this prompt the target label are not provided. Intuitively, the \emph{非上述类型} label is the most difficult one since it has to confidently rule out all the given labels. The confusion matrix in Figure \ref{fig:confusion_matrix} shows that LLMs have high F1 scores on all 10 real categories, while it performs the worst on the \emph{非上述类型} label with 60.2\% F1. Most mistakenly classified \emph{非上述类型} samples are assigned the \emph{注意事项} (precautions) or \emph{功效作用} (effects) labels. Table \ref{tab:error_qic} reports three error cases on the KUAKE-QIC task.

\begin{table}[ht]
\caption{Error cases of the fine-tuned BaiChuan 13b on the KUAKE-QIC task.  }
\label{tab:error_qic}
\begin{center}
\begin{tabular}{p{9cm}p{1.5cm}p{1.5cm}}
\hline
\textbf{Input Sentence}  &   \textbf{Ground truth}  & \textbf{Predicted}  \\
\hline

请识别以下搜索query的意图类型：$\backslash$n臀位怎么睡才能转头位图片$\backslash$n搜索意图选项：功效作用，病因分析，医疗费用，治疗方案，指标解读，注意事项，后果表述 \quad (Please identify the intent type of the following search query: $\backslash$nHow to sleep to turn breech position to head position picture$\backslash$nSearch intent options: efficacy, etiology analysis, medical expenses, treatment options, indicator interpretation, precautions , statement of consequences)   &  非上述类型 (not in the given categories)  &   注意事项 (precautions)   \\
\hline 

请根据提供的搜索词，选择最符合其意图的类型：$\backslash$n天麻枸杞三七可泡酒吗$\backslash$n意图选项：注意事项，就医建议，功效作用，疾病描述，指标解读，治疗方案，病因分析，后果表述 \quad (Please select the type that best suits your intention based on the search terms provided: $\backslash$nCan Gastrodia, Lycium Barbarum and Panax Notoginseng be soaked in wine?$\backslash$nIntention options: Precautions, medical advice, efficacy and effect, disease description, indicator interpretation, treatment plan, etiology analysis, consequences statement)    &  非上述类型 (not in the given categories)  &   功效作用 (efficacy and effect)   \\

\hline 

如何快速减肥$\backslash$n根据搜索关键词，回答这个搜索的意图是什么？$\backslash$n类型选项：注意事项，就医建议，病因分析，功效作用，指标解读，治疗方案，医疗费用，病情诊断 \quad (How to lose weight quickly$\backslash$nBased on the search keywords, what is the purpose of this search? $\backslash$nType options: precautions, medical advice, etiology analysis, efficacy and effect, indicator interpretation, treatment plan, medical expenses, disease diagnosis)  &  治疗方案 (treatment plan)   &  非上述类型 (not in the given categories)   \\

\hline 

\end{tabular}
\end{center}
\end{table}

On the test set of the IMCS-V2-MRG task, BaiChuan 13b's generated reports for patient-doctor dialogues have an average of 102.35 Chinese tokens, compared to 105.08 Chinese tokens in the ground-truth reports written by doctors. BaiChuan 13b achieves 47.6 ROUGE-L score. Table \ref{tab:error_mrg} reports one example that achieves the lowest ROUGE-L score (13.1) on the test set. Despite that the dialogue mentioned the patient (User's child) demonstrated elevated platelets after having coughs, the finetuned LLM fails to mention this information in the predicted history of present illness. And it fails to summarize the medical examination results in the dialogue to the auxiliary examinations section. We believe the presented example demonstrates that the open-sourced Chinese LLMs still have room for improvements in terms of long medical text understanding, which is the most challenging part of the IMCS-V2-MRG task.

\begin{table}[ht]
\caption{A case of LLM for the IMCS-V2-MRG task. Since the patient-doctor dialogue is long, we just put a special token [dialogue] here, and the complete conversation and its English translation can be found in Appendix \ref{subsec:example_dialogue}. }
\label{tab:error_mrg}
\begin{center}
\begin{tabular}{p{2cm}p{11cm}}
\hline
\textbf{Prompt}  &   根据下面医患对话，为患者生成一份综合性的诊疗报告。包含主诉、现病史、辅助检查、既往史、诊断，以及建议等六个部分。$\backslash$n问诊对话历史：$\backslash$n[\emph{dialogue}] \quad (Based on the following doctor-patient dialogue, a comprehensive diagnosis and treatment report is generated for the patient. It includes six parts: chief complaint, history of present illness, auxiliary examinations, past history, diagnosis, and suggestions. $\backslash$nConsultation conversation history:$\backslash$n[dialogue] )  \\
\hline

\multicolumn{2}{c}{\textbf{主诉 (chief complaint)}} \\
\hdashline
Ground truth  &   咳嗽 (coughing)     \\
Predicted   &      血小板升高 (elevated platelets)  \\
\hline

\multicolumn{2}{c}{\textbf{现病史 (history of present illness)}} \\
\hdashline
Ground truth  &   4个月大患儿出现咳嗽，就诊提示白细胞及血小板增多 (A 4-month-old child developed a cough, and the doctor showed an increase in white blood cells and platelets.)     \\
Predicted   &      患儿无明显诱因下出现血小板升高情况，无咳嗽咳痰，无恶心呕吐，无其他明显不适症状。精神状态一般，胃纳一般，余如常。 (The child had elevated platelets without obvious inducement, had no cough or sputum, no nausea and vomiting, and no other obvious symptoms of discomfort. The mental state is normal, the appetite is normal, and everything else is as usual.)  \\
\hline 

\multicolumn{2}{c}{\textbf{辅助检查 (auxiliary examinations)}} \\
\hdashline
Ground truth  &   血常规提示白细胞增多、血小板增多 (Routine blood test showed increased white blood cells and increased platelets)     \\
Predicted   &      暂无 (None) \\
\hline

\multicolumn{2}{c}{\textbf{既往史 (past history)}} \\
\hdashline
Ground truth  &   无 (None)     \\
Predicted   &      既往体健。 (Previously in good health.)  \\
\hline

\multicolumn{2}{c}{\textbf{诊断 (diagnosis)}} \\
\hdashline
Ground truth  &   小儿上呼吸道感染 (upper respiratory tract infection in children)     \\
Predicted   &      小儿感冒 (Children's cold)  \\
\hline 

\multicolumn{2}{c}{\textbf{建议 (suggestions)}} \\
\hdashline
Ground truth  &   控制感染治疗；服用三七或者阿司匹林或者是潘生丁治疗 (Infection control treatment; take Panax notoginseng or aspirin or dipyridamole for treatment)     \\
Predicted   &    继续当前治疗，密观。 (Continue the current treatment and esoteric meditation.)  \\
\hline

\hline

\hline 

\end{tabular}
\end{center}
\end{table}

\subsubsection{How does LLMs perform under few-shot fine-tuning setting?}

This ranking is different from the few-shot fine-tuning setting, demonstrating that we need to use different PEFT methods under different training settings.

Under the few-shot fine-tuning setting, we use the training data used as demonstrations in Table \ref{tab:results_few_shot_in_context} as the available training samples and use this set of 122 samples for fine-tuning the open-sourced LLMs, ChatMed and BaiChuan 13b. Table \ref{tab:results_few_shot_ft} reports the results of few-shot fune-tuning with P-tuning-v2, Adapter and LoRA. Note that for the convenience of presentations, we will just report the average scores on each cohort of tasks on PromptCBLUE. From the table \ref{tab:results_few_shot_ft}, we can observe that: (a) With the few-shot training set, the fine-tuned BaiChuan 13b under-performs the three powerful LLM APIs in Table \ref{tab:results_few_shot_in_context}. (b) Undertandably, the few-shot fine-tuning performance is much worse compared to Table \ref{tab:results_ft}, where the LLMs are fine-tuned with a large-scale training set. (c) among the parameter efficient tuning methods, the overall performance ranking is: P-tuning-v2 $>$ LoRA $>$ Adapter, which is slightly different from the ranking under the large scale fine-tuning setting. The results demonstrate that we need to use different PEFT methods under different training settings.

\begin{table}[t]
\caption{The few-shot fine-tuning results on the PromptCBLUE test set. In this experiment, we use the demonstration samples from the previous experiment (Table \ref{tab:results_few_shot_in_context}) as the training set. }
\label{tab:results_few_shot_ft}
\begin{center}
\begin{tabular}{c|ccc}
\hline
\textbf{Task}   &   P-tuning-v2   & Adapter  &   LoRA     \\
\hline
\multicolumn{4}{c}{\textbf{\emph{Results of Chinese-LlaMA-2 7B}}}  \\
\hline
Medical information extraction  &  0.238   &  0.219  &  0.224    \\
Medical text classification  &   0.263  &  0.226  &  0.251   \\
Medical natural language inference  &   0.314   &   0.295   &  0.309  \\
Symptom status understanding  &   0.331   &   0.316   &  0.308  \\
Medical content generation  &   0.135  &  0.142   &  0.127   \\
\hline
Overall score   &   0.260   &    0.242   &   0.249   \\
\hline 

\multicolumn{4}{c}{\textbf{\emph{Results of Chinese-LlaMA-2 13B}}}  \\
\hline
Medical information extraction  &  0.287   &  0.264  &  0.278    \\
Medical text classification  &   0.336  &  0.338  &  0.327   \\
Medical natural language inference  &   0.352   &   0.328   &  0.351  \\
Symptom status understanding  &   0.364   &   0.359   &  0.357  \\
Medical content generation  &   0.153  &  0.145   &  0.142   \\
\hline
Overall score   &   0.305   &    0.291   &   0.298   \\
\hline 

\end{tabular}
\end{center}
\end{table}

\section{Conclusion}

In this paper, we present the PromptCBLUE benchmark, which re-purpose the CBLUE benchmark to the LLM era. We carefully transfer all the traditional medical natural language processing tasks into the unified prompt-response generation form. Our PromptCBLUE benchmark provide a good testbed for Chinese LLMs to test their capabilities in the medical domain and their potentials for various medical applications. This paper also conduct extensive experiments and case studies to analyze how the current open-sourced Chinese LLMs perform on PromptCBLUE.

\section{Acknowledgements}

We would like thank the following experts in the field of computer science and medicine for serving as the advising committee members of the PromptCBLUE benchmark\footnote{The committee members are ranked alphabetically. }: (a) Liang Chen, from Huashan Hospital of Fudan University. (b) Liang He, from East China Normal University. (c) Xuanjing Huang, from Fudan University. (d) Haofen Wang, from Tongji University. (e) Xiaochun Yang, from Northeastern University.

\bibliography{iclr2024_conference}
\bibliographystyle{iclr2024_conference}

\appendix

\section{Appendix}

\subsection{Prompt and target formats for PromptCBLUE}
\label{subsec:format}

We now present one example of the prompt and response formats for each sub-task of PromptCBLUE. For complete set of prompt templates, please refer to our project page \url{https://github.com/michael-wzhu/PromptCBLUE}.

\noindent\textbf{CMeEE-v2} \quad As a standard named entity recognition task in the medical domain, when given a input text sequence, one has to extract the medical entities, including the text mentions’ spans and entity types, in the sequence. Recently, with the prevailing of generative LM, a series of literature no longer requires the model to predict the entity spans. Thus, we formulate CMeEE-v2 task as the task of generating all the entity mentions under given entity types. 

Prompt Template:
\begin{verbatim} 
医学实体识别: (Medical named entity recognition: )
[sentence]
实体选项: [entity_types] (Choices of entity types: [entity_types])
答: (Answer: )
\end{verbatim}

Target Template:
\begin{verbatim} 
上述句子中的实体包含: (The above sentence have the following entities: )
[entity_type]实体: [entity_mentions] ([entity_type] entities: 
[entity_mentions])
\end{verbatim}

Example prompt: 
\begin{verbatim}
医学实体识别: (Medical named entity recognition: )
外周血白细胞计数常明显升高，伴核左移。 (The peripheral blood white blood cell 
count is often significantly elevated, with left nuclear shift.)
实体选项: 疾病，医学检验项目，医院科室，身体部位，微生物类，临床表现，药物 (Choices of entity types: diseases, medical test items, hospital 
departments, body parts, microorganisms, clinical manifestations, 
drugs)
答： (Answer: )
\end{verbatim}

Example target: 
\begin{verbatim} 
上述句子中的实体包含: (The above sentence have the following entities: )
医学检验项目实体: 外周血白细胞计数 (medical test items entities: 
peripheral blood white blood cell count)
疾病实体: 无 (disease entities: None)
医院科室实体: 无 (hospital department entities: None)
药物实体: 无 (drug entities: None)
微生物类实体: 无 (microorganism entities: None)
\end{verbatim}

Note that we also want LLM to determine whether there are entities for a given type in the sentence.

\noindent\textbf{CMeIE} \quad The CMeIE task is a triple extraction task, that is, the model has to predict the head entity information (including mention span and entity type), tail entity information (including mention
span and entity type), and the relation type between the two entities. Similar as the CMeEE task, we reformuate this task as the task of generating the head-tail entity pair under the given relation types. The following are the prompt and target templates, in which [relations] denotes the relation names, and [sentence] denotes the input text. An example pair of prompt and target response is also presented. 

Prompt Template:
\begin{verbatim} 
找出句子中的具有[relations]的头尾实体对: (Find the entity pairs that 
have the [relations] relations: )
[sentence]
\end{verbatim}

Target Template:
\begin{verbatim} 
上述句子中[relation]关系的实体对如下: [triples] (the entity pairs having 
the [relation] relation are as follows: [triples])
上述句子中[relation]关系的实体对如下: [triples] (the entity pairs having 
the [relation] relation are as follows: [triples])
\end{verbatim}

Example prompt: 
\begin{verbatim}
找出句子中的具有临床表现，同义词关系类型的头尾实体对： (Find the entity pairs that have the clinical finding or 
synonym relations:)
妊娠期高血压@健康孕妇可以通过全身血管阻力 (systemic vascular resistance，
SVR) 和心输出量下降得以平衡，而妊娠期高血压妇女SVR较低，心输出量往往会增加。 (Hypertension during pregnancy@healthy pregnant women can be 
balanced by a decrease in systemic vascular resistance (SVR) and 
cardiac output, whereas women with gestational hypertension have a 
lower SVR and tend to have an increase in cardiac output.)
\end{verbatim}

Example target: 
\begin{verbatim} 
上述句子中临床表现关系的实体对如下： (the entity pairs having 
the clinical finding relation are as follows: )
头实体：妊娠期高血压，尾实体：SVR较低；头实体：妊娠期高血压，尾实体：心输出量往往会增加； (head entity: hypertension during pregnancy, tail entity: SVR 
is lower; head entity: hypertension during pregnancy, tail entity: 
cardiac output tends to increase;)
上述句子中同义词关系的实体对如下： (the entity pairs having the synonym 
relation are as follows: )
无 (None)
\end{verbatim}

\noindent\textbf{CHIP-CDEE} \quad In this task, one needs to extract clinical findings as medical events in a given medical report. An clinical finding event consists of a 主体词 (trigger), and three attributes, 发生状态 (occurrence status), 描述词 (descriptor) and 解剖部位 (
anatomical part). In PromptCBLUE, this task is reformulated as generating event descriptions for clinical findings given a medical report. 

Prompt Template:
\begin{verbatim} 
[sentence]
问题：句子中的临床发现事件及其属性是什么？ (Question: What are the clinical 
findings and their attributes in the sentence?)
说明：临床发现事件由主体词，发生状态，描述词和解剖部位组成 (Note: a clinical finding event consists of trigger, occurrence 
status, descriptor and anatomical part.)
\end{verbatim}

Target Template:
\begin{verbatim} 
上述句子中的临床发现事件如下： (The clinical finding events in the above 
sentence are as follows:)
主体词：[str]；发生状态：[str]；描述词：[str]；解剖部位：[str] (trigger: [str]; occurrence status: [str]; descriptor: 
[str]; anatomical part: [str])
\end{verbatim}

Example prompt: 
\begin{verbatim}
7月前患者给予亚砷酸氯化钠(伊泰达)注射液 10mg 静滴14天，6月前予以口服维甲酸 20mg bid*14天维持治疗，5月前行亚砷酸氯化钠(伊泰达)注射液 10mg 静滴14天维持化疗，3月余前复查骨髓检查示增生性骨髓象；fish：pml/rara（双色双融合）(15/17)：未见异常；腰穿脑脊液未见异常细胞。现为维持化疗入院。(Before July, patients were given 10mg of sodium arsenite chloride 
(Itada) injection intravenously for 14 days. Before June, they were 
given oral retinoic acid 20mg bid * 14 days for maintenance treatment. 
Before May, patients were given 10mg of sodium arsenite chloride 
(Itada) injection intravenously for 14 days for maintenance 
chemotherapy. After more than 3 months, bone marrow examination 
showed proliferative myelogram; Fish: pml/rara (dual color dual 
fusion) (15/17): no abnormalities found; No abnormal cells were 
found in the cerebrospinal fluid through lumbar puncture. I am 
currently admitted for maintenance chemotherapy.)
问题：句子中的临床发现事件及其属性是什么？ (Question: What are the clinical 
findings and their attributes in the sentence?)
说明：临床发现事件由主体词，发生状态，描述词和解剖部位组成 (Note: a clinical finding event consists of trigger, occurrence 
status, descriptor and anatomical part.)
\end{verbatim}

Example target: 
\begin{verbatim}
上述句子中的临床发现事件如下： (The clinical finding events in the above 
sentence are as follows:)
主体词：fish：pml/rara（双色双融合）(15/17)异常；发生状态：否定；描述词：无；解剖部位：无 (trigger: Fish: pml/rara (dual color dual fusion) (15/17) anomaly; 
occurrence status: not found; descriptor: none; anatomical part: none)
主体词：骨髓象；发生状态：无；描述词：增生性；解剖部位：骨髓 (trigger: bone marrow imaging; occurrence status: none; descriptor: 
proliferative; anatomic site: bone marrow)
\end{verbatim}

\noindent\textbf{CHIP-CDN} \quad This task aims to map the diagnosis terms (query terms) written by doctors to standardized disease terms (target terms) according to a certain medical terminology standard \cite{jmir}. The number of standardized terms may exceed 10 thousands. The standard system adopted in this task is usually the International Statistical Classification of Diseases and Related Health Problems 10th Revision (ICD-10)\footnote{https://www.who.int/standards/classifications/classification-of-diseases}. ICD-10 has more than 30 thousand disease terms, thus it is prohibitive to feed all the disease terms into ChatGPT. ICD-10 has different versions in different countries, and in this work we adopt the ICD-10 Beijing Clinical Trial (version v601)\footnote{http://www.cips-chip.org.cn/2021/eval3}. We will refer to this Chinese version of ICD-10 as ICD-10-Beijing. 

Recently, this task is usually addressed by a retrieve-and-rank system \cite{Park2019AnIR}. Firstly, a small set of candidate standard terms are retrieved via inverted index (e.g., ElasticSearch) or a semantic index toolkit like Faiss, then a ranking model will score and rank the relevances of each query-target term pair. To construct the data samples in CBLUE, for each query term in the task datasets, we first retrieve candidate target terms using BM25 \cite{Robertson2009ThePR}. The candidate pool consists of the top 10-30 terms retrieved by BM25 not are not in the ground truth, and the ground truth terms (for 80\% of the cases).  The candidate pool's order is shuffled so that LLMs can not obtain the true answers by just selecting the first few candidates. Thus, under PromptCBLUE, LLMs act as a ranking model, and choose the final target terms among the candidates or determine no proper target terms are present in the candidate pool.

Prompt Template:
\begin{verbatim} 
[sentence]
术语选项：[candidate-terms] (choices of terms：[candidate-terms])
说明：从候选的若干个ICD-10诊断标准词中选择出与原诊断描述匹配的词 (Task explanation: select the terms that match the original 
diagnostic description from several candidate ICD-10 diagnostic 
standard terms)
答： (Answer)
\end{verbatim}

Target template:
\begin{verbatim} 
[selected-terms]
\end{verbatim}

Example prompt: 
\begin{verbatim}
主动脉弓缩窄心功能低下 (Aortic arch constriction and cardiac dysfunction)
术语选项：胫前动脉假性动脉瘤，主动脉缩窄，男性性腺功能低下，男性性腺功能低下，垂体功能低下，心功能不全 (choices of terms：pseudoaneurysm of the anterior tibial artery, 
coarctation of the aorta, male hypogonadism, male hypogonadism, 
pituitary dysfunction, and cardiac insufficiency)
说明：从候选的若干个ICD-10诊断标准词中选择出与原诊断描述匹配的词 (Task explanation: select the terms that match the original 
diagnostic description from several candidate ICD-10 diagnostic 
standard terms)
答： (Answer)
\end{verbatim}

Example target: 
\begin{verbatim}
主动脉缩窄，心功能不全 (coarctation of the aorta, cardiac insufficiency)
\end{verbatim}

\noindent\textbf{CHIP-CTC} \quad The main objective of this task is to classify the text input into a type of clinical trial screening criteria (44 types). Under PromptCBLUE, we transform the task to the task of generate the type of clinical trial screening criteria for the give text input and the given label set. LLMs need to respond  with "Not the above type" to indicate that the given label types in the prompt does not match the sentence.

Prompt template:
\begin{verbatim} 
[sentence]
这句话是什么临床试验筛选标准类型？ (What is the type of clinical trial 
screening criteria for this sentence?)
类型选项：[candidate-types] (Choices of categories: [candidate-types])
\end{verbatim}

Target template:
\begin{verbatim} 
[selected-type]
\end{verbatim}

Example prompt: 
\begin{verbatim}
8.过去3个月内有过眼内手术的患者； (8. Patients who have undergone intraocular 
surgery within the past 3 months;)
这句话是什么临床试验筛选标准类型？ (What is the type of clinical trial 
screening criteria for this sentence?)
类型选项：成瘾行为，吸烟状况，性取向，残疾群体，读写能力，肿瘤进展，参与其它试验，疾病分期，能力，疾病，药物，诊断，教育情况，口腔相关，受体状态，健康群体，数据可及性，设备，献血，过敏耐受，特殊病人特征，睡眠，怀孕相关，研究者决定，器官组织状态，症状(患者感受)，治疗或手术，护理，性别，种族，实验室检查，知情同意，饮食，年龄，居住情况，病例来源，酒精使用，体征(医生检测），锻炼，风险评估，预期寿命，伦理审查，依存性 (Choices of categories: addiction behavior, smoking status, sexual 
orientation, disabled population, reading and writing ability, tumor 
progression, participation in other trials, disease staging, ability, 
disease, medication, diagnosis, educational status, oral related, 
receptor status, healthy population, data accessibility, equipment, 
blood donation, allergic tolerance, special patient characteristics, 
sleep, pregnancy related, researcher decision, organ and tissue status, 
symptoms (patient perception), treatment or surgery, Nursing, gender, 
race, laboratory examination, informed consent, diet, age, residential 
status, source of case, alcohol use, physical signs (doctor tested), 
exercise, risk assessment, life expectancy, ethical review, dependency)
\end{verbatim}

Example target: 
\begin{verbatim}
治疗或手术 (treatment or surgery)
\end{verbatim}

\noindent\textbf{KUAKE-QIC} \quad This task asks one to classify the intent labels for an online medical search query. There is a relatively high proportion of samples with the the "other" category. Thus, similar to the KUAKE-QIC task, we drop this label and reformulate this task as generating the intent type of a given search query with the candidate labels given in the prompt, or respond with "Not the above type" to indicate that the given label types in the prompt does not match the sentence.

Prompt Template:
\begin{verbatim} 
判断下面搜索词的意图： (Determine the intention of the following 
search term)
[query] 
选项：[candidate-types] (Choices: [candidate-types])
答： (Answer: )
\end{verbatim}

Target template:
\begin{verbatim} 
[selected-type]
\end{verbatim}

Example prompt: 
\begin{verbatim}
判断下面搜索词的意图： (Determine the intention of the following search 
term)
武汉传染性尖锐湿疣的治疗方法 (Treatment methods for infectious genital 
warts in Wuhan)
选项：指标解读，治疗方案，功效作用，注意事项，病情诊断，就医建议，疾病描述 (Choices: Indicator interpretation, treatment plan, efficacy, 
precautions, condition diagnosis, medical advice, disease description)
答： (Answer: )
\end{verbatim}

Example target: 
\begin{verbatim}
治疗方案 (treatment plan)
\end{verbatim}

\noindent\textbf{CHIP-STS} \quad The aim of this task is to determine whether two disease-related questions express the same semantics. Under PromptCBLUE, the task is to respond 是 (\emph{yes}), or 相同 (the same), or "不是" (\emph{no}), or 不同 (different) to indicate whether the two input medical questions in the prompt have the same semantics.

Prompt Template:
\begin{verbatim} 
下面两个句子语义是“相同”或“不同”？ (Are the following two sentences 
semantically the same or different?)
“[sentence-1]”，“[sentence-2]”。
选项：[candidate-types]
答：(Answer: )
\end{verbatim}

Target template:
\begin{verbatim} 
[selected-type]
\end{verbatim}

Example prompt: 
\begin{verbatim}
下面两个句子语义是“相同”或“不同”？ (Are the following two sentences 
semantically the same or different?)
“糖尿病的三多一少是什么”，“无限极的“灵芝皇”和“桑唐饮”能治好糖尿病吗？”。 ("What is diabetes' three more and one less ", "Can the 
limitless" Ganoderma lucidum emperor "and" Sangtang drink "cure 
diabetes?".)
选项：相同，不同 (Option: the ame, different)
答：(Answer: )
\end{verbatim}

Example target: 
\begin{verbatim}
不同 (different)
\end{verbatim}

\noindent\textbf{KUAKE-QTR} \quad Under CBLUE, this task asks one to determine to what degree the online medical search query and the title of a web page are related semantically by outputing a relevance score. The relevance score is one of the followng four values: 0 (means "完全不匹配", \emph{completely mismatched or without any reference value}), 1 (means "很少匹配有一些参考价值", \emph{slightly matched but having some reference value}), 2 (means "部分匹配", \emph{partially matched}) and 3 (means "完全匹配", \emph{perfectly matched}). Under PromptCBLUE, this task is formulated as outputing the label names indicating to what degree a query and a page title are related. 

Prompt Template:
\begin{verbatim} 
下面的搜索词和页面标签的意思有多相同？ (How do the following search terms 
and page tags have the same meaning?)
搜索词：[query] (Search term: [query])
页面标题：[web-page-title] (Page Title: [web-page-title])
选项：完全不匹配或者没有参考价值，很少匹配有一些参考价值，部分匹配，完全匹配 (Option: Completely mismatched or without reference value, rarely 
matched with some reference value, partially matched, completely 
matched)
\end{verbatim}

Target template:
\begin{verbatim} 
[selected-type]
\end{verbatim}

Example prompt: 
\begin{verbatim}
下面的搜索词和页面标签的意思有多相同？ (How do the following search terms 
and page tags have the same meaning?)
搜索词：宝宝三周了发烧不玩睡觉 (Search term: The baby has been running 
a fever for three weeks, he doesn't play and just sleep)
页面标题：孩子三周了手足口发烧一天就不烧了就是睡觉打搀 (Page Title: The child has been running a fever in his hands, feet, 
and mouth for three weeks, but after a day, he stops burning and 
goes to bed)
选项：完全不匹配或者没有参考价值，很少匹配有一些参考价值，部分匹配，完全匹配 (Option: Completely mismatched or without reference value, rarely 
matched with some reference value, partially matched, completely 
matched)
\end{verbatim}

Example target: 
\begin{verbatim}
部分匹配 (partially matched)
\end{verbatim}

\noindent\textbf{KUAKE-QQR} \quad This task is designated to differentiate the semantic relations between two medical queries. And the semantic relation labels are: -1 (meaning "语义无直接关联", \emph{not semantically related}), 0 (meaning "后者是前者的语义父集", \emph{the latter is the semantic superset of the former}), 1 (meaning "后者是前者的语义子集", \emph{the latter is the semantic subset of the former}), 2 (meaning "完全一致", \emph{completely equivalent}).

Prompt template:
\begin{verbatim} 
下面两个句子的语义关系是？ (What is the semantic relationship between the 
following two sentences?)
“[sentence-1]”，“[sentence-2]”。 
选项: 完全一致，后者是前者的语义子集，后者是前者的语义父集，语义毫无关联 (Option: completely consistent, the latter is a semantic subset 
of the former, and the latter is a semantic superset of the former, 
with no semantic correlation)
\end{verbatim}

Target template:
\begin{verbatim} 
[selected-type]
\end{verbatim}

Example prompt: 
\begin{verbatim}
下面两个句子的语义关系是？ (What is the semantic relationship between 
the following two sentences?)
“伤口涂什么药好得快”，“有伤口涂什么药”。 (What kind of medicine is good 
for wounds quickly? What kind of medicine is good for wounds.) 
选项: 完全一致，后者是前者的语义子集，后者是前者的语义父集，语义毫无关联 (Option: completely consistent, the latter is a semantic subset 
of the former, and the latter is a semantic superset of the former, 
with no semantic correlation)
\end{verbatim}

Example target: 
\begin{verbatim}
完全一致 (completely consistent)
\end{verbatim}

\noindent\textbf{KUAKE-IR} \quad This is originally a standard information retrieval (IR) task in the medical domain. It asks one to retrieve relevant documents for a online medical query among a corpus of 1000 thousand documents. Apparently, one can not input all the documents in the corpus into LLMs. Thus, we reformulate this task to the task of determine whether the given query and document are relevant. 

Prompt template:
\begin{verbatim} 
医疗搜索：[query] (medical search: [query])
回答内容：[document] (content: [document])
选项: 相关，不相关 (Option: related, unrelated)
答： (Answer: )
\end{verbatim}

Target template:
\begin{verbatim} 
[selected-type]
\end{verbatim}

Example prompt: 
\begin{verbatim}
医疗搜索：鼻梁被撞鼻梁矫正手术 (medical search: correction surgery for 
nasal bridge collision)
回答内容：你好，你这中情况一般需要行鼻骨截骨整形及鼻中隔联合矫正，手术需要住院，大概需要10天左右的时间，费用在12000左右，我们医院不对医保，如果是要医保报销需要办转诊手续之后再凭相关单据回所在地报销。 (content: Hello, in this case, you usually need to undergo nasal 
bone osteotomy and nasal septum joint correction. The surgery requires 
hospitalization, which takes about 10 days and costs around 12000 
yuan. Our hospital does not provide medical insurance. If you want 
medical insurance reimbursement, you need to go through referral 
procedures and then return to your location with relevant documents 
for reimbursement.)
选项: 相关，不相关 (Option: related, unrelated)
答： (Answer: )
\end{verbatim}

Example target: 
\begin{verbatim}
相关 (unrelated)
\end{verbatim}

\noindent\textbf{CHIP-MDCFNPC} \quad Given a dialogue history between a patient and a doctor, this task asks one to extract the clinical finding entities and their statuses. Here the status of a clinical finding entity is defined as one of the four categories: "阳性" ("已有症状疾病或者假设未来可能发生的疾病", \emph{existing symptoms/diseases or potential future symptoms/diseases}), "阴性" (meaning "未患有的症状疾病", \emph{symptoms/diseases not currently experienced}), 其他 ("没有回答、不知道、回答不明确或者模棱两可不好推断", \emph{no answer, don't know, unclear answer, or ambiguous response that is not conducive to inference.}), 不标注 ("无实际意义的不标注或者和病人当前的状态独立不标注", \emph{irrelevant and not specified or unrelated to the patient's current condition and not specified}). Under PromptCBLUE, we ask LLMs to output the descriptions of the status label for extracted clinical finding entities.

Prompt template:
\begin{verbatim} 
[dialogue-history]
问题：上述问诊对话中临床发现有哪些？这些实体的阴阳性是？ (Question: What are the clinical findings in the above consultation 
dialogue? What are the statuses of these entities?)
阴阳性选项：已有症状疾病或者假设未来可能发生的疾病等，未患有症状疾病，没有回答、不知道、回答不明确或者模棱两可不好推断，无实际意义的不标注或者和病人当前的状态独立不标注 (Options: Existing symptomatic diseases or assumed future diseases, 
etc., without symptomatic diseases, without answers, unclear answers, 
or ambiguous reasoning, without practical significance, or independent 
of the patient's current state)
说明：临床发现是临床医学下，病人状态描述的概念集合 (Explanation: Clinical findings are a collection of concepts 
describing patient states in clinical medicine)
\end{verbatim}

Target template:
\begin{verbatim} 
上述对话中临床发现实体以及其阴阳性判别如下：(The clinical finding entities detected in the above conversation 
and its status are as follows: )
[clinical-finding-entity]：[status-option]
\end{verbatim}

Example prompt: 
\begin{verbatim}
患者：月经来了还可吃乌鸡白凤丸和丹栀逍遥丸吗 (Patient: Can I still take Wuji Baifeng 
Pills and Danzhi Xiaoyao Pills after menstruation)
医生：请问类似症状出现多长时间？ (Doctor: How long has similar symptoms 
occurred?)
医生：你吃这药是治疗什么的 (Doctor: What is the purpose of taking this 
medicine for)
患者：我前几个月去检查是游离子腺素增高，月经没来 (Patient: I went to check a few months ago for an increase in free 
radicals, and my menstrual cycle did not come)
医生：什么高 (Doctor: what is high?)
患者：甲状游离子腺素增高 (Patient: Increased thyroid free radical hormone)
医生：把化验单给我看一下。 (Doctor: Show me the test report.)
患者：现在去检查正常值了 (Patient: Now we're going to check the normal values)
医生：那你如果月经量多这些药就不吃了，如果月经量少就可以吃。 (Doctor: If you have a lot of menstrual flow, you can skip 
these medications. If you have a little menstrual flow, you can take them.)
患者：就是月经不调 (Patient: It's just menstrual irregularities)
患者：甲状腺素药还有吃 (Patient: Still taking thyroid hormone medication)
医生：是甲状腺功能低下吗？甲减吗？ (Doctor: Is it hypothyroidism? Is 
hypothyroidism present?)
患者：我在马来西亚看不懂报告单 (Patient: I cannot understand the report 
form in Malaysia)
医生：嗯嗯，只有甲状腺功能低下才需要吃甲状腺素。 (Doctor: Hmm, only hypothyroidism requires taking thyroid hormone.)
问题：上述问诊对话中临床发现有哪些？这些实体的阴阳性是？ (Question: What are the clinical findings in the above consultation 
dialogue? What are the statuses of these entities?)
阴阳性选项：已有症状疾病或者假设未来可能发生的疾病等，未患有症状疾病，没有回答、不知道、回答不明确或者模棱两可不好推断，无实际意义的不标注或者和病人当前的状态独立不标注 (Options: Existing symptomatic diseases or assumed future diseases, 
etc., without symptomatic diseases, without answers, unclear answers, 
or ambiguous reasoning, without practical significance, or independent 
of the patient's current state)
说明：临床发现是临床医学下，病人状态描述的概念集合 (Explanation: Clinical findings are a collection of concepts describing 
patient states in clinical medicine)
\end{verbatim}

Example target: 
\begin{verbatim}
上述对话中临床发现实体以及其阴阳性判别如下：(The clinical finding entities detected in the above conversation 
and its status are as follows: )
月经没来：已有症状疾病或者假设未来可能发生的疾病等 (Menstruation not coming: Existing symptomatic diseases or assumed 
future diseases, etc.)
游离子腺素增高：已有症状疾病或者假设未来可能发生的疾病等 (Elevated free radical adenosine: Existing symptomatic diseases 
or assumed future diseases, etc.)
甲状游离子腺素增高：已有症状疾病或者假设未来可能发生的疾病等 (Increased thyroid free radical hormone: Existing symptomatic 
diseases or assumed future diseases, etc.)
月经量少：无实际意义的不标注或者和病人当前的状态独立不标注 (Low menstrual flow: No labeling of meaningless or independent of 
the patient's current state)
月经量多：无实际意义的不标注或者和病人当前的状态独立不标注 (Menorrhagia: Low menstrual flow: No labeling of meaningless or 
independent of the patient's current state)
月经不调：已有症状疾病或者假设未来可能发生的疾病等 (Irregular menstruation: Existing symptomatic diseases or assumed 
future diseases, etc.)
甲减：没有回答、不知道、回答不明确或者模棱两可不好推断 (Hypothyroidism: Not answering, not knowing, unclear or ambiguous 
in answer, making it difficult to infer)
甲状腺功能低下：无实际意义的不标注或者和病人当前的状态独立不标注 (Low menstrual flow: No labeling of meaningless or independent of 
the patient's current state)
\end{verbatim}

\noindent\textbf{IMCS-V2-SR} \quad Under CBLUE, this task consists of multiple steps: (a) extract symptom entities from a dialogue history between a patient and a doctor; (b) normalize the symptom entity mentions to standardized terms (choosing from the vocabulary provided by the task); (c) determine the status of a symptom entity. In this task, the status label set is: "患有该症状" (\emph{experiencing/suffering from the symptom}), "没有患有该症状" (\emph{not experiencing/suffering from that symptom}), "无法根据上下文确定病人是否患有该症状" (unable to determine from the context whether the patient is experiencing that symptom). Under PromptCBLUE, we discard the second step, that is, we only ask the LLMs to extract the symptom entities in the current utterance and determine their statuses.

Prompt template:
\begin{verbatim} 
找出当前对话中的症状，并判断阴阳性： (Identify the symptoms in the current 
conversation and determine their status:)
对话历史：(Dialogue history: )
[dialogue-history]
当前对话：(Current utterance: )
[current-utterance]
症状阴阳性选项：没有患有该症状，患有该症状，无法根据上下文确定病人是否患有该症状 (Options of symptom status: not experiencing/suffering from that 
symptom, experiencing/suffering from the symptom, unable to determine 
from the context whether the patient is experiencing that symptom)
\end{verbatim}

Target template:
\begin{verbatim} 
当前对话中的症状及其阴阳性判断为：(The symptoms in the current conversation 
and their statuses are: )
[aymptom]：[status]
\end{verbatim}

Example prompt: 
\begin{verbatim}
找出当前对话中的症状，并判断阴阳性： (Identify the symptoms in the current 
conversation and determine their status:)
对话历史： (Dialogue history: )
患者：没有怎么听啊 (Patient: I didn't listen very much)
医生：根据您的描述，宝宝咳嗽，嗓子吼，可能是气喘或喉鸣，考虑支气管炎的可能性较大 (Doctor: According to your description, the baby is coughing, 
roaring in the throat, possibly due to wheezing or wheezing in the 
throat, and the possibility of bronchitis is higher)
当前对话： (Current utterance: )
医生：需要带宝宝去医院儿科就诊，用听诊器听诊肺部，查血常规胸片等相关检查，排除肺炎，根据结果，给于控制感染，止咳化痰等对症治疗。 (Doctor: It is necessary to take the baby to the pediatric 
department of the hospital for treatment, use a stethoscope to 
auscultate the lungs, conduct blood routine chest X-ray and other 
related tests, rule out pneumonia, and provide symptomatic treatment 
such as infection control, cough relief, and phlegm reduction 
based on the results.)
症状阴阳性选项：没有患有该症状，患有该症状，无法根据上下文确定病人是否患有该症状 (Options of symptom status: not experiencing/suffering from that 
symptom, experiencing/suffering from the symptom, unable to determine 
from the context whether the patient is experiencing that symptom)
\end{verbatim}

Example target: 
\begin{verbatim}
当前对话中的症状及其阴阳性判断为：(The symptoms in the current conversation 
and their statuses are: )
肺炎：无法根据上下文确定病人是否患有该症状 (Pneumonia: unable to determine from the context whether the patient
is experiencing that symptom)
感染：患有该症状 (Infection: experiencing/suffering from the symptom)
咳：患有该症状 (Cough: experiencing/suffering from the symptom)
痰：患有该症状 (Phlegm: experiencing/suffering from the symptom)
\end{verbatim}

\noindent\textbf{IMCS-V2-NER} \quad The reformulation of this task is similar to that of CMeEE-v2, except that this task focuses on NER of dialogue utterances. 

Prompt template:
\begin{verbatim} 
下面对话中的[entity-labels]实体有哪些？ (What are the [entity-labels] 
entities in the following conversation?)
[utterance]
答： (Answer: )
\end{verbatim}

Target template:
\begin{verbatim} 
上述句子中的实体包含： (The entities in the above sentence include: )
[entity-type]实体：[entity-mentions]
\end{verbatim}

Example prompt: 
\begin{verbatim}
下面对话中的医学检查检验，症状，医疗操作实体有哪些？ (What are the medical examinations, symptoms, and medical procedures 
in the following conversation?)
宝贝也呕吐吗？ (Does the baby also vomit?)
答： (Answer: )
\end{verbatim}

Example target: 
\begin{verbatim}
上述句子中的实体包含： (The entities in the above sentence include: )
医学检查检验实体：无 (Medical examination and testing entity: None)
症状实体：呕吐 (Symptom entity: vomiting)
医疗操作实体：无 (Medical operation entity: None)
\end{verbatim}

\noindent\textbf{IMCS-V2-DAC} \quad This task askes one to classify the intent of the current utterance in a patient-doctor dialogue for online medical consultations. The original task only gives label names that are not natural languages. To better fit this task into LLMs, we re-write the intent label names. 

Prompt template:
\begin{verbatim} 
确定这句话的意图: (Determine the intent label of this utterance: )
[utterance]
类型选项：[candidate-labels] (Choices of intent labels: [candidate-labels])
\end{verbatim}

Target template:
\begin{verbatim} 
[selected-label]
\end{verbatim}

Example prompt: 
\begin{verbatim}
确定这句话的意图: (Determine the intent label of this utterance: )
当时医生说我们单纯支气管炎也不喘就开的药 (At that time, the doctor said we 
prescribed medication for simple bronchitis without wheezing)
类型选项：关于就医建议的解答，给出诊断，关于症状的回答，关于症状的询问，关于就医建议的提问，关于已有检查和治疗的回答，关于注意事项的提问，关于已有检查和治疗的提问，关于个人基本信息的询问，关于个人基本信息的回答，关于用药建议的解答，关于病因的询问，关于用药建议的提问，关于注意事项的解答，关于病因的回答 (Choices of intent labels: Answers to medical advice, providing 
diagnosis, answering symptoms, asking questions about symptoms, 
asking questions about medical advice, answering questions about 
existing tests and treatments, asking questions about precautions, 
asking questions about existing tests and treatments, asking questions 
about personal basic information, answering questions about personal 
basic information, answering questions about medication advice, 
asking questions about causes, and asking questions about medication 
advice, Answers to precautions and etiology)
\end{verbatim}

Example target: 
\begin{verbatim}
关于已有检查和治疗的回答 (answers regarding existing examinations and 
treatments)
\end{verbatim}

\noindent\textbf{IMCS-V2-MRG} \quad In this task, LLMs are asked to summarize a patient-doctor dialogue and write a diagnostic and treatment report. In the dataset, all the reports are divided into 6 sections: chief complaint, present illness history, auxiliary examination, past medical history, diagnosis, recommendations. Thus, under PromptCBLUE, we also ask the LLMs to generate the summarization report in the order of these six sections.

Prompt template:
\begin{verbatim} 
问诊对话历史： (Consultation conversation history: )
[dialogue-history]
根据上述对话，给出诊疗报告 (Based on the above dialogue, a diagnosis and 
treatment report is given)
说明：诊疗报告分为主诉, 现病史, 辅助检查, 既往史, 诊断, 建议这六个章节。 (Note: The diagnosis and treatment report is divided into six 
chapters: chief complaint, history of present illness, auxiliary 
examination, past history, diagnosis, and recommendations.)
\end{verbatim}

Target template:
\begin{verbatim} 
上述问诊对话的诊疗报告如下： (The diagnosis and treatment report of the 
above consultation dialogue is as follows:)
主诉：[str] (Chief complaint: [str])
现病史：[str] (History of current illness: [str])
辅助检查：[str] (Auxiliary check: [str])
既往史：[str] (Past history: [str])
诊断：[str] (Diagnosis: [str])
建议：[str] (Suggestion: [str])
\end{verbatim}

Example prompt: 
\begin{verbatim}
问诊对话历史： (Consultation conversation history: )
患者：宝宝刚满月，母乳喂养，最近两天时不时的会咳嗽一声，食欲和精神还行，只不过睡觉不是很安稳。家里面最近两天大人和宝宝的姐姐也有感冒，不知道宝宝是被传染了感冒还是怎么样，请问怎么治疗？ (Patient: The baby is just one month old and breastfeeding. He 
has coughed from time to time in the past two days. His appetite 
and energy are okay, but his sleep is not very stable. In the past 
two days at home, the adults and the baby's sister have also had colds. 
I don't know if the baby has been infected with a cold or something 
else. How can I treat it?)
医生：您好，我是您的辅诊医生，需要询问几个问题，才能更好的评估孩子情况，您还在吗？ (Doctor: Hello, I am your auxiliary doctor. I need to ask a few 
questions to better evaluate the child's condition. Are you still here?)
医生：宝宝体温正常吗？ (Doctor: Is the baby’s temperature normal?)
医生：还在吗？ (Doctor: Are you still there?)
患者：在 (Patient: in)
医生：您好 (Doctor: Hello)
医生：宝宝现在体温正常吗 (Doctor: Is the baby’s temperature normal now?)
患者：体温正常 (Patient: Temperature is normal)
医生：口吐泡泡吗 (Doctor: Are you spitting bubbles at the mouth?)
患者：没有 (Patient: None)
医生：嗓子哑吗 (Doctor: Do you have a hoarse voice?)
患者：哭起来跟以前一样 (Patient: Crying is the same as before)
医生：好的 (Doctor: OK)
患者：只不过鼻音重 (Patient: It’s just that the nasal sound is heavy)
医生：还有其他症状吗 (Doctor: Are there any other symptoms?)
患者：睡不安稳 (Patient: Unable to sleep well)
医生：出汗多吗？ (Doctor: Do you sweat a lot?)
患者：不多 (Patients: not many)
医生：哭闹吗 (Doctor: Are you crying?)
患者：比以前爱哭闹 (Patient: more crying than before)
医生：大便什么样子 (Doctor: What does the stool look like?)
患者：这个没注意，昨天一天没有大便，今天上午大便的，大便以后睡得安稳一些了，不过还是时不时咳嗽一声 (Patient: I didn’t pay attention to this. He didn’t have a bowel 
movement yesterday. He had a bowel movement this morning. He slept 
more peacefully after the bowel movement, but he still coughed from 
time to time.)
医生：有痰吗? (Doctor: Is there phlegm?)
患者：没有痰，干咳 (Patient: There is no phlegm, and there is dry cough)
医生：嗯嗯 (Doctor: Uh-huh)
医生：流鼻涕吗 (Doctor: Do you have a runny nose?)
患者：没有 (Patient: None)
医生：嗯嗯 (Doctor: Uh-huh)
医生：因为宝宝比较小，最好带宝宝去公立医院儿科就诊，听一下肺部没有问题的话可以观察看看 (Doctor: Because the baby is relatively small, it is best to take 
the baby to the pediatrician in a public hospital. If there is no 
problem with the lungs, he can be observed.)
根据上述对话，给出诊疗报告 (Based on the above dialogue, a diagnosis 
and treatment report is given)
说明：诊疗报告分为主诉, 现病史, 辅助检查, 既往史, 诊断, 建议这六个章节。 (Note: The diagnosis and treatment report is divided into six 
chapters: chief complaint, history of present illness, auxiliary 
examination, past history, diagnosis, and recommendations.)
\end{verbatim}

Example target: 
\begin{verbatim}
上述问诊对话的诊疗报告如下： (The diagnosis and treatment report of the 
above consultation dialogue is as follows:)
主诉：阵发性咳嗽。 (Chief complaint: 
Paroxysmal cough.)
现病史：患儿阵发性干咳两天。 (History of current illness: The child had 
paroxysmal dry cough for two days.)
辅助检查：暂缺。 (Auxiliary check: None)
既往史：不详。 (Past history: None)
诊断：咳嗽待查。 (Diagnosis: Cough to be examined.)
建议：儿科就诊，听诊肺部。 (Suggestion: Pediatric visit, auscultation of 
the lungs.)
\end{verbatim}

\noindent\textbf{MedDG} \quad This task asks one to generate the responses of doctors give the dialogue history and the patient's current utterance. Under the original setting, MedDG is an entity-centric dialogue generation task, that is, it asks the model to generate responses that contains certain entities. Under PromptCBLUE, we currently discard this setting and only asks the LLMs to generate responses on behlf of the doctors. 

Prompt template:
\begin{verbatim} 
[dialogue-history]
根据上述对话历史，作为医生应该如何回复？ (Based on the above conversation 
history, how should a doctor respond?)
答： (Answer: )
\end{verbatim}

Target template:
\begin{verbatim} 
[generated-response]
\end{verbatim}

Example prompt: 
\begin{verbatim}
患者：最近总是到了晚上就胃很难受。这几天吃过饭就有点反胃但是都是头痛头晕恶心。胃还是很难受（女，19岁） (Patient: Recently, I always feel sick in my stomach at 
night. I've been feeling a little nauseous after eating for the past 
few days, but I still have headaches, dizziness, and nausea. My 
stomach still feels uncomfortable (female, 19 years old))
医生：你好，这种情况有多长时间了？ (Doctor: Hello, how long has this been 
going on?)
患者：半个月了。 (Patient: It’s been half a month.)
医生：平时吃饭规律吗？ (Doctor: Do you eat regularly?)
根据上述对话历史，作为医生应该如何回复？ (Based on the above conversation 
history, how should a doctor respond?)
答： (Answer: )
\end{verbatim}

Example target: 
\begin{verbatim}
胃部感觉难受是怎么难受？反酸烧心打嗝？还是胃疼胃胀？ (Why does your stomach feel so bad? Acid reflux, heartburn and 
hiccups? Or stomachache or bloating?)
\end{verbatim}

\subsection{Data set statistics for the PromptCBLUE}
\label{subsec:stat}

In the Table \ref{tab:stats}, we present the dataset statistics for PromptCBLUE.

\begin{table}[t]
\caption{Dataset statistics for the PromptCBLUE benchmark. }
\label{tab:stats}
\begin{center}
\begin{tabular}{c|ccccc}
\hline
\textbf{Task}   &   \#Train   & \#dev  &   \#test    &  Avg Prompt length    &   Avg target length      \\
\hline
\multicolumn{6}{c}{\textbf{\emph{Sub-tasks}}}  \\
\hline

CMeEE-V2  &  5000   &  400   &  400    &   107.88   &   54.03   \\
CMeIE   &  5000   &  400   &  400    &    293.72   &  135.51  \\
CHIP-CDEE   &  3000   &  400    &  400   &  142.61   &  180.93    \\
CHIP-CDN   &  5000   &  400    &  400    &   281.79   &  10.37    \\
CHIP-CTC   &   6600   &  704   &  704    &   214.61   &    3.81   \\
CHIP-STS   &   5000   &  400    &  400    &   66.26    &   2.0   \\
KUAKE-QIC   &   5500    &  440    &  440    &  81.58   & 
   4.09   \\
KUAKE-QTR   &  5000    &  400   &  400    &    96.38   & 
  7.23   \\
KUAKE-QQR   &  5000  &  400    &  400    &   89.38    &  7.61   \\

KUAKE-IR   &    5000   &  400   &  400    &     203.33   &    2.78    \\
CHIP-MDCFNPC  &   5000   &  400    &  400    &   744.99  
 &   67.67   \\
IMCS-V2-SR  &   5000   &  400    &  400    &   137.13   &    36.33   \\
IMCS-V2-NER   &   5000   &  400    &  400    &  61.66  & 
  23.65   \\
IMCS-V2-DAC    &   5000    &  512   &  512    &     371.62   &     8.56   \\
IMCS-V2-MRG   &    3000    &  400    &  400     &    821.1   &  105.08   \\
MedDG    &   5000   &  400    &  400    &   194.75   &  27.71  \\ 
\hline
\multicolumn{6}{c}{\textbf{\emph{Total}}}  \\
\hline
PromptCBLUE   &    82600    &    7656    &    7656     &    265.22   &  71.10    \\
\hline

\end{tabular}
\end{center}
\end{table}

\subsection{Hyper-parameter settings for the PEFT methods}
\label{subsec:hyperparams}

We now elaborate on the hyper-parameter settings for different PEFT methods, and the training settings.

\textbf{PEFT Hyper-parameters} \quad For P-tuning, we prepend 128 soft prompt tokens to the input. The learnable prompts are randomly initialized and use the LSTM prompt encoder. For P-tuning-V2, each layer's soft prompt is directly randomly initialized without prompt encoder. The prompt length is also 128 for P-tuning-V2. Adapter is added in parallel to the feed-forward and self-attention module of the LLM, with a bottleneck dimension of 128. LoRA's rank is set to 24 and is added to the query, key, value, output weight matrix, and the two weight matrix in the feed-forward module. AdaLoRA's initial rank setting is the same with LoRA.

\textbf{Training settings} \quad Each PEFT method is trained with the same training hyper-parameters: warm up steps are 100, batch size is 64, and the learning rate is 3e-4, the optimizer is AdamW, the learning schedule is linear decay, gradient clipping is 1, weight decay is 1e-4, and the maximum epoch is 10.  The other settings follow the huggingface Transformers' default settings. We find that training a few steps after the dev loss hits the lowest point is helpful for generating better responses. Thus we pick the checkpoint that is 200 steps after the one with the lowest dev loss.

\subsection{Example dialogue for Table \ref{tab:error_mrg}}
\label{subsec:example_dialogue}

The complete patient-doctor conversation for Table \ref{tab:error_mrg} is as follows:
\begin{verbatim} 
患者：四个月宝宝血小板701有问题吗？宝宝最近有点咳嗽，去医院查血，白细胞十点几，血小板701，医生说怕是血栓。宝宝有点贫血，还有点缺钙，可能是因为奶水不是很多。昨天我感冒了，不知道有没有传染给宝宝。血小板高还可能是血液病，听一些网上的医生说这样的情况不用干预，会自己好的？？请专业医生解答一下，宝宝的问题大吗？应该怎么办呢？
医生：你好
患者：你好
患者：血小板高有问题吗？
医生：这个孩子的血小板偏高，应该与感染有一定的关系
医生：病毒感染或者说是炎症很容易导致血小板升高
医生：这个血小板数值达不到血小板增高症的诊断标准，一般来说超过八百，甚至说九百，才能诊断为血小板增高症
患者：有可能是血栓吗？
医生：血小板偏高会导致血液的粘滞度增加，发生血栓的概率比正常的孩子大的多，但是这种概率比较低
患者：问题大吗？
医生：我看你提供的病史，妈妈有感冒的情况，应该考虑被妈妈传染了
医生：血液病的可能性不大，应该考虑感染有一定的关系
医生：一般来说，这个血小板数值是不需要特殊处理的，随着孩子感染的控制，血小板的数值会逐渐的下降
患者：那怎么办？
医生：给孩子看感染治疗就可以了，因为这个孩子白细胞总数偏高啊，感染控制以后复查血常规
患者：问题不大吗
医生：是的
患者：嗯，谢谢！
医生：你好
患者：曲大夫你好
医生：三七可以吃个，具有活血化瘀的作用
患者：孩子今晚拉肚子了，看来炎症在肠胃。
医生：应该考虑胃肠道的
患者：可是才四个月大，能吃吗
医生：中药制剂的咳嗽药物可能会导致孩子出现拉肚子
医生：不建议口服的，这个孩子太小了
医生：如果就像口服的话，可以选择阿司匹林或者是潘生丁，这两个药物具有抑制血小板聚集的作用
患者：哦，孩子有点贫血
医生：血红蛋白是多少
患者：还没复查呢
医生：哦
患者：嗯，谢谢医生！
医生：这个孩子最主要的问题，是感染所导致的血小板升高，白细胞升高
\end{verbatim}

And its English translation is:
\begin{verbatim} 
Patient: Is there any problem with the platelet 701 of the four-month-old 
baby? The baby had a cough recently. I went to the hospital for a 
blood test. The white blood cells were over 10 and the platelets 
were 701. The doctor said it was probably a blood clot. The baby 
is a little anemic and a little calcium deficient, maybe because 
there is not a lot of milk. I caught a cold yesterday and I don’t 
know if it was passed on to my baby. High platelets may also be a 
blood disease. I heard some doctors on the Internet say that this 
kind of situation will heal on its own without intervention? ? 
Please ask a professional doctor to answer. Is the baby’s problem 
serious? How should I do it?
Doctor: Hello
Patient: Hello
Patient: Is there a problem with high platelets? 
Doctor: This child's platelets are high, which should be related 
to infection
Doctor: Viral infection or inflammation can easily lead to 
elevated platelets
Doctor: This platelet value does not meet the diagnostic criteria 
for thrombocytosis. , generally speaking, it can be diagnosed as 
thrombocytosis if it exceeds 800, or even 900.
Patient: Is it possible to have a blood clot? 
Doctor: High platelets will increase blood viscosity, and the 
probability of thrombosis is much higher than that of normal 
children, but this probability is relatively low
Patient: Is it a big problem? 
Doctor: From the medical history you provided, my mother has a 
cold, so I should consider that she was infected by her
Doctor: The possibility of blood disease is unlikely, and I should 
consider that infection is related to it.
Doctor: Generally speaking, this platelet value does not require 
special treatment. As the child's infection is controlled, the 
platelet value will gradually decrease
Patient: What should I do? 
Doctor: Just treat the child for infection, because the total number 
of white blood cells in this child is high. After the infection is 
controlled, the blood routine will be rechecked
Patient: Is it not a big problem?
Doctor: Yes
Patient: Well, thank you! 
Doctor: Hello
Patient: Hello, Dr. Qu
Doctor: You can eat Panax notoginseng, which has the effect of 
promoting blood circulation and removing blood stasis
Patient: My child has diarrhea tonight. It seems that the 
inflammation is in the intestines and stomach. 
Doctor: Gastrointestinal problems should be considered
Patient: But he is only four months old, can he take it?
Doctor: Traditional Chinese medicine preparations for cough medicine 
may cause diarrhea in children
Doctor: It is not recommended to take it orally because the child 
is too young
Doctor: If it is taken orally, you can choose aspirin or dipyridamole. 
These two drugs have inhibitory effects. The role of platelet 
aggregation
Patient: Oh, the child is a little anemic
Doctor: What is the hemoglobin
Patient: I haven’t reviewed it yet
Doctor: Oh
Patient: Well, thank you doctor! 
Doctor: The main problem with this child is the increase in platelets 
and white blood cells caused by infection.
\end{verbatim}

\end{CJK*}

\end{document}